\documentclass[hf]{ceurart}

\usepackage{listings}
\newcommand{\circled}[1]{\raisebox{.5pt}{\textcircled{\raisebox{-.9pt} {\small #1}}}}

\definecolor{lightyellow}{HTML}{fef9e7} %

\newcommand{\ML}{{ML}}
\newcommand{\LLMs}{{LLMs}}
\newcommand{\LLM}{LLM}

\newcommand{\sensitive}{{protected}} 
\newcommand{\groundtruth}{{ground truth}} %

\newcommand{\decisionmaking}{{decision-making}}

\definecolor{asparagus}{rgb}{0.53, 0.66, 0.42}
\definecolor{bittersweet}{rgb}{1.0, 0.44, 0.37}
\definecolor{chocolate}{rgb}{0.82, 0.41, 0.12}
\definecolor{cadmiumgreen}{rgb}{0.0, 0.42, 0.24}
\definecolor{ao(english)}{rgb}{0.0, 0.5, 0.0}

\renewcommand{\mid}{\, | \,}

\newcommand{\filledownarrow}{%
    \tikz[baseline=-0.1ex]{%
        \draw[-{Latex[length=1.8mm,width=1.8mm]}, fill=black] (0,0.25) -- (0,-0.1);%
    }%
}

\newcommand{\timespan}{%
    \tikz[baseline=-0.1ex]{%
        \draw[|-|] (-0.1,0.1) -- (0.1,0.1);%
    }%
}

\usepackage{amssymb}

\usepackage[utf8]{inputenc} 
\usepackage[T1]{fontenc}    
\usepackage{url}            
\usepackage{booktabs}       
\usepackage{amsfonts}       
\usepackage{nicefrac}       
\usepackage{microtype}      
\usepackage{xcolor}         
\usepackage{amsthm}         

\usepackage{mathtools}
\usepackage{xspace}
\usepackage{bbm}
\usepackage{pifont}
\usepackage{caption}
\usepackage{subcaption}
\usepackage{graphicx}
\usepackage{wrapfig}
\usepackage{enumitem}
\usepackage{multirow}
\usepackage{comment}
\usepackage{setspace}
\usepackage{multicol}
\usepackage{algorithm}
\usepackage{amsmath}

\usepackage{tikz}
\usetikzlibrary{bayesnet}
\usetikzlibrary{arrows}
\usetikzlibrary{shapes,decorations,arrows,calc,arrows.meta,fit,positioning}
\tikzset{
    -Latex, auto, node distance = 0.5 cm and 0.5 cm, semithick,
    state/.style = {circle, draw, minimum width = 0.7 cm},
    const/.style = {minimum width = 0.7 cm},
    inter/.style = {rectangle, draw, minimum width = 0.7 cm, minimum height = 0.7 cm},
    point/.style = {circle, draw, inner sep = 0.04cm, fill, node contents = {}},
    bidirected/.style = {Latex-Latex,dashed},
    el/.style = {inner sep=2pt, align=left, sloped}
}

\usepackage{mdframed}

\usepackage{makecell} 
\usepackage{tabularx}

\usepackage[normalem]{ulem}

\begin{document}

\ifcasreviewlayout
    \linenumbers
\else
\fi

\copyrightyear{2022}
\copyrightclause{Copyright for this paper by its authors. Use permitted under Creative Commons License Attribution 4.0 International (CC BY 4.0).}

\conference{EWAF'24: European Workshop on Algorithmic Fairness, July 01--03, 2024, Mainz, Germany}

\title{Algorithmic Fairness Over Time: Advances \& Prospects}

\author{Miriam Rateike}[
email=rateike@cs.uni-saarland.de
]
\address{Saarland University,
 Saarbrücken, Germany}
\begin{abstract}
In the presence of feedback loops, algorithmic decisions at one time step can influence observed data in subsequent steps, potentially introducing selection bias or shifts in data distribution. Research has demonstrated that ignoring these effects can lead to unfairness, which may worsen over time as the model remains deployed within society. This raises important questions of how to measure and ensure fairness of algorithmic decision-making models over time. In this short paper, we present a taxonomy that distinguishes between different types of static and temporal fairness notions in recent advancements. We then suggest prospects for future research by identifying common implicit temporal assumptions, open questions to enhance our temporal understanding of fairness, and other open challenges.
\end{abstract}


\maketitle

\section{Introduction}

In many contexts such as loan approval~\citep{liu2018delayed}, pretrial release decisions~\cite{lum2016statistical}, or fraud detection~\cite{kumar2019credit}, decisions have consequences over time~\citep{kilbertus2021beyond}. %
They often influence either the data collected, or the underlying data distribution, or both. 
This is due to \emph{feedback loops}, where algorithmic decisions at one time step influence data in subsequent time steps and consequently impact future decisions~\citep{ chaney2018algorithmic, fuster2022predictably, pagan2023classification}. 
Supervised classification models in such scenarios often focus on accurate predictions and do not consider the \emph{impact of decision} on data collection or the underlying distribution during deployment~\citep{kilbertus2021beyond, perdomo2020performative, scantamburlo2023prediction}. Consequently, algorithmic fairness has primarily been studied in a static environment, assuming an unchanging data generation process~\citep{hardt2016equality, dwork2012fairness, agarwal2018reductions, zafar2017fairness}.

As algorithms are increasingly being deploying in society, neglecting the downstream effects of their decisions can be detrimental; it can result in suboptimal performance of learned (fair) \ML\ models~\citep{liu2018delayed, kilbertus2020fair, damour2020fairness}, can perpetuate or amplify existing biases~\citep{liu2018delayed,  damour2020fairness, creager2020causal, Hu_Zhang_2022, paassen2019dynamic}, or introduce new biases~\citep{ghassami2020characterizing, sachs2005causal, strobl2019constraint, vowels2021d}.
As a result, there has been a call to study the \emph{temporal dimensions} of fairness~\cite{chouldechova2018frontiers, schwobel2022long, bohdal2023fairness}. However, a comprehensive and contextual survey of existing work on temporal fairness, including future prospects---beyond a recent technical review~\citep{gupta2023temporal}---is lacking.

In this short paper, our contribution is twofold: we introduce a taxonomy of recent advancements in fairness of algorithmic systems deployed in society in the presence of feedback loops (\S~\ref{sec:method1}), and we propose prospects and open challenges to guide future research (\S~\ref{sec:method2}).
Here, we focus on fairness in algorithmic decision-making using tabular data.
A longer version of this paper will expand sections, include temporal fairness in ranking, and provide a more contextualized discussion.
We start with a brief background of the problem setting and resulting challenges.

\paragraph{Problem Setting.}
Consider a lending scenario where a bank seeks to deploy an algorithmic policy to decide on loan applications. 
Let an individual in a population be described by some sensitive features $s$ (e.g., gender), non-sensitive features $x$ (e.g., income), and a ground truth $y$ (e.g., repayment ability) following a ground truth distribution
\mbox{$p_{true} = p(y | x, s)p(x, s)$}. 
 At each time step, the algorithmic policy determines whether to approve or deny credit to applicants based on their characteristics.
The policy takes as input an individual's observed features $(x, s)$ and outputs a probability distribution over decisions $d$, i.e.,  $p_d = p(d \mid x, s)$. 
The bank seeks to train the policy to accurately predict the unobserved \groundtruth\ $y$ aiming to provide credit only to individuals capable of repayment.

In a typical classification machine learning (\ML) pipeline, we first sample data from the ground truth distribution. Then, we train the ML model on the training data and deploy it on test data in the subsequent step.
The majority of work on fair classification~\citep{hardt2016equality, dwork2012fairness, agarwal2018reductions,zafar2017bfairness, zafar2019fairness} assumes that i) observed (train and test) data are independently and identically distributed (i.i.d.) samples from the ground truth population ($p_{obs} = p_{true}$); and ii) training and test data are sampled from the same distribution ($p_{test} = p_{train}$). 
When algorithms are deployed in society using real-world data, these assumptions may not hold due to different types of feedback loops~\citep{pagan2023classification}, which lead to different challenges.

\paragraph{Challenges.}
First, in the real world, often only a particular decision $d$ reveals \groundtruth\ $y$. For example, we only know if someone would have paid back a loan, if we actually granted the person a loan, but not if we rejected their loan application. 
The \groundtruth\ (labels) only materializes after a decision has been granted.
This introduces a bias known as \emph{selection bias}, where the observed data is not an i.i.d. sample from the population of interest~\citep{liu2018delayed, lakkaraju2017selective}. 
For example, if a policy at $t_1$ only grants individuals from one gender identity loans, then we observe in $t_2$ the \groundtruth\ only for that gender identity, but we have no \groundtruth\ information about other gender identities in the population.  
The observed data in $t_2$ is not an i.i.d. sample from the ground truth distribution $p_{true}$, but biased by the decision policy $p_{obs} \propto p_d \, p_{true}$, such that $p_{obs} \neq p_{true}$~\citep{kilbertus2020fair}.
When the observed data is used in the next step, such as for retraining the \ML\ model, a \emph{sampling feedback loop}~\citep{pagan2023classification} from decisions to the data sampling process exists~\citep{Rateike2022DontTI, kilbertus2017avoiding, bechavod2019equal}.

Second, when deploying a policy in the real world, a decision at time $t_1$ may alter the data generative process in $t_2$ (sometimes referred to as \emph{performative prediction}~\citep{perdomo2020performative}). 
For example, individuals may respond to a negative decision in $t_1$ (e.g., denial of a loan) by changing the values of their features in $t_2$ (e.g., paying off a previous loan)~\citep{liu2018delayed, damour2020fairness, creager2020causal}. 
This means that the population undergoes a distribution shift.
When testing an algorithm learned on data at time $t_1$ on data at time $t_2$, the test data is no longer drawn from the same distribution as the training data, i.e., $p_{\text{test}} \neq p_{\text{train}}$.
In this case feedback loops from decisions to the data generative process exist, causing shifts in the data distribution.
Thereby, decisions can affect unobserved or observed features or the \groundtruth\ via \emph{individual}, \emph{feature}, or \emph{outcome feedback loops}, respectively~\citep{pagan2023classification}.
Firms that deploy algorithmic systems can thus potentially use them to steer populations towards more desirable behaviors, exerting \emph{performative power}~\citep{hardt2022performative}. 
In the fairness context, a firm with high predictive power can reinforce existing societal norms and stereotypes---or facilitate transformative changes towards a more equitable society.\footnote{Whether the transformative power should be leveraged and under which framework remains an open question, and we call interdisciplinary research to engage in this discussion, already initiated by~\citep{schwobel2022long, wachter2020bias, susser2022decision}.} 

Note that both challenges can also co-exist. Selection biases poses a particular challenge when biased data is used to retrain the model, creating an \emph{ML model feedback loop}~\citep{pagan2023classification};
 when data distributions shift unexpectedly in response to decisions, retraining becomes essential.

\section{Advances: Temporal Dimensions of Fairness}\label{sec:method1}

\begin{wrapfigure}{r}[2pt]{0.17\textwidth}
  \vspace{-10pt}
  \centering
  \begin{tikzpicture}[node distance=0.2cm]

    \node(node2) {};
    \node[right=0.15 of node2] (node2c) {};
     \node[right=0.27 of node2] (node2d) {};
    \node[right=0.8 of node2] (right2) {};
     \node[below=0.05 of right2] (right2low) {};
    \node[right=0.65 of right2] (node2b) {};
    \node[right=0.8 of right2] (node2a) {}; 

    \node[right=0.01 of node2] (node2tick1) {};
    \node[right=0.65 of node2] (node2tick2) {};
    
    \node[right=0.645 of node2] (node2tick3) {};

     \node[right=1.3 of node2] (node2tick4) {};

    \node[above=0.02 of node2] (node3) {};
    \node[right=0.8 of node3] (right3) {};
    \node[right=0.8 of right3] (node3a) {}; 
    \node[above=0.3 of node3] (node4) {};
    \node[right=0.8 of node4] (right4) {};
    \node[right=0.7 of right4] (node4b) {};
    \node[right=0.8 of right4] (node4a) {}; 
    \node[above=0.1 of node4] (node5) {};
    \node[right=0.8 of node5] (right5) {};
    \node[right=0.8 of right5] (node5a) {}; %
    \node[right=0.7 of right5] (node5b) {};
    \node[right=0.2 of node5] (node5c) {};


    \node[above=0.15 of node2d](node1c) {\circled{5}};
    \node[above=0.1 of node5](node1) {};
    \node[above=0.3 of right4] (right1) {\circled{1}};
    \node[above=0.3 of node4b] (node1b) {\circled{2}
    };

    \draw[->] (right1) -- (right4);
    \draw[->] (node1b) -- (node4b);

    \draw[->, line width=1.5pt, arrows = -{Stealth[length=5mm]}] (node2) -- (node2a);
    \draw[|-|, line width=1pt] (node2tick1) -- (node2tick2);
    \draw[|-|, line width=1pt] (node2tick3) -- (node2tick4);

    \draw[|-|] (node3) -- (node3a) ;
    \draw[|-|] (right4) -- (node4a) node[midway, above=0.01mm]{\circled{4}};
    \draw[|-|] (right4) -- (node4) node[midway, above=0.01mm]{\circled{3}};

    \draw[-, dashed] (right2low) -- (right1);
    \draw[-] (right2) -- (right2) node[midway, below=1mm]{$\pi_t$};
    \draw[-] (node2c) -- (node2c) node[below=0.05mm]{{\scriptsize{time}}};
  
\end{tikzpicture}%
  \caption{}
  \label{fig:1}
\end{wrapfigure}

When algorithms are deployed in society over longer periods of time, temporal dimensions of fairness become important (Fig.~\ref{fig:1}).
Traditional fairness criteria~\citep{ hardt2016equality, dwork2012fairness} 
do not explicitly incorporate time into their definition of fairness. They measure similarities between individuals~\citep{dwork2012fairness}, or group-dependent statistics~\citep{hardt2016equality, dwork2012fairness} at a specific moment in \mbox{time (\filledownarrow)}.
We refer to them as \emph{non-temporal fairness notions}.
They can be asked to hold at time $t$ a decision is made with policy $\pi_t$ (\circled{1} \emph{immediate fairness}), or to be fulfilled only in a distant point in time (\mbox{\circled{2} \emph{long-term fairness}}). 
Instead, \emph{temporal fairness notions} explicitly model time in their definition, typically by defining a duration over which fairness needs to hold (\timespan). 
This duration can lie in the past ({\circled{3} \emph{fairness in hindsight}}) or in the future (\mbox{\circled{4} \emph{fairness in foresight}}) of current decision-making, or both (\circled{5} \emph{fairness across time}) and can be finite or infinite. 

\paragraph{Non-temporal Fairness Notions.}
\emph{Immediately fair} (\circled{1}) or myopic policies aim for non-temporal fairness at each time step of decision making~\citep{damour2020fairness, zhang2020fair}.
Many works on fair sequential decision making take this approach~\citep{damour2020fairness, wen2021algorithms}, often under sampling feedback loops~\citep{kilbertus2020fair, Rateike2022DontTI, bechavod2019equal}.
Immediate fairness can contribute to a rapid building of trust in the adoption of an algorithmic policy.
However, immediately fair policies can increase inequalities of a population's feature or qualification distributions over time, especially when feedback loops induce distribution shifts~\citep{damour2020fairness, creager2020causal, zhang2020fair}.
\emph{Long-term fairness} (\circled{2}) instead aims for non-temporal fairness criteria to hold in a distant future~\cite{yin2023long, rateike2023designing, puranikdynamic, yu2022policy, weber2022enforcing}.
Long-term fair approaches often leverage an algorithmic system's performative power and shift data distributions to mitigate systemic biases~\citep{yin2023long, rateike2023designing}, which can align with EU non-discrimination principles~\citep{wachter2020bias}.
In this case, temporal effects are often modeled during training, e.g., by relying on reinforcement learning approaches.
Experimental results show that optimizing for long-term fairness can eventually yield higher fairness and utility than short-term approaches~\citep{paassen2019dynamic, rateike2023designing}, but often come at the cost of immediate fairness needs~\citep{yin2023long, rateike2023designing}.
Long-term fair approaches often rely on strong assumptions about the dynamical system, such as time homogeneity~\citep{wen2021algorithms, yin2023long, rateike2023designing}.
If assumptions fail, it may yield suboptimal or unfair outcomes.

\paragraph{Temporal Fairness Notions.}
Temporal fairness monitors algorithmic systems over time, and asks for similar individuals to be treated similarly at different points in time~\citep{Hu_Zhang_2022, gupta2021individual,  heidari2018preventing} or policies to obtain similar fairness levels at different points in time~\citep{Rateike2022DontTI}.
Thereby the fairness of a policy $\pi_t$ is assessed by comparing its decisions to past and/or future decisions. 
\emph{Fairness across time} (\circled{5}) demands fairness to be independent of time, which holds for past and future decisions~\citep{heidari2018preventing, gupta2021individual, wen2021algorithms}. For example, a temporal version of individual fairness~\cite{dwork2012fairness} asks for similar individuals from any time step to receive similar treatment~\citep{gupta2021individual}. 
\emph{Fairness in hindsight} (\circled{3}) evaluates fairness with respect to the history of decisions~\citep{gupta2021individual, zhao2023fairo, alamdari2023remembering}, e.g., asking individuals to receive comparable or improved decisions today compared to similar individuals from the past~\citep{gupta2021individual}. 
Different from this retrospective view, \emph{fairness in foresight} (\circled{4}) focuses the future impact of an algorithmic policy~\citep{Hu_Zhang_2022, Rateike2022DontTI, zezulka2023fairness}.
Examples are a temporal version of path-specific counterfactual fairness~\citep{chiappa2019path} considering causal paths over time~\citep{Hu_Zhang_2022}, 
and a notion of temporal variance, requiring consistent group fairness of policies over time~\citep{Rateike2022DontTI}.
The time dependency of these criteria can be relaxed, by requiring weaker fairness the further decisions are apart~\citep{gupta2021individual}, or by limiting the duration to a shorter time frame~\citep{heidari2018preventing}.

\section{Prospects}\label{sec:method2}

\paragraph{Beyond Traditional Assumptions.}
Existing approaches to sequential algorithmic \decisionmaking\ often rely on implicit assumptions, which may not align with the real-world or may lose validity over time. 
As such, it becomes important to explore more realistic scenarios.
For example, prior work often assumes \emph{static \sensitive\ attributes} defined by legally protected concepts like race or gender usually modeled as root nodes in causal or generative models~\cite{damour2020fairness, zhang2020fair, wen2021algorithms, yin2023long, rateike2023designing}. However, \sensitive\ attributes are often societal constructs, which can change over time~\citep{hu2020sex, sen2016race}.
Further, algorithms may unintentionally create new \sensitive\ groups over time~\cite{wachter2022theory} that are legally unprotected~\citep{kirchhefer2023algorithmicgroups}.
Existing approaches often also assume that responses to decisions (e.g., payment behavior following a granted loan) are \emph{observed immediately}, and that feedback effects (e.g., changes in credit score following repayment) \emph{occur immediately}~\citep{ wen2021algorithms, yin2023long, puranikdynamic, yu2022policy, xu2023equal, chi2022towards, jabbari2017fairness}. However, in practice, these mechanisms can take years (e.g., recidivism is reported in 2-year intervals~\citep{COMPAS}), and updating a policy every few years can significantly slow down iterative learning approaches. Other assumptions include the absence of \emph{feedback loops between features}, challenged by \citep{hu2020sex, mhasawade2021causal}, and that decisions on \emph{one individual do not influence others}, differently~\citep{heidari2019effort}, and the availablility of large amounts of data to learn the dynamical system~\citep{yin2023long, weber2022enforcing, xu2023equal, chi2022towards}.

\paragraph{Advancing Fairness Considerations.}
There are several prospects of advancing the discussion of temporal dimensions of fairness. 
First, we require a more thorough investigation of the \emph{trade-offs between immediate and long-term fairness} found in prior work~\citep{yin2023long, rateike2023designing}, as prioritizing immediate fairness may overlook systemic biases that perpetuate over time~\citep{damour2020fairness, wachter2020bias}, while long-term fairness may lead to unfairness in the short term~\citep{yin2023long}.
Second, existing work often enforces a single fairness metric over time, yet \emph{societal fairness needs are dynamic}, requiring more flexible \ML\ frameworks. 
Third, dynamical systems often involve \emph{multiple stakeholders}, including decision-makers, individuals reacting to decisions, and government enforcing regulations~\citep{scantamburlo2023prediction, schwobel2022long, schumann2020we}. We thus need to develop frameworks for context-aware fairness that incorporate different stakeholder needs.
Fourth, the role of \emph{societal interventions} beyond decision-making has been repeatedly mentioned in the literature~\citep{ mhasawade2021causal, heidari2019effort, von2020fairness}, yet it remains understudied.

\paragraph{Open Challenges in Dynamical Systems.}
We identify several open challenges that affect system dynamics and consequently algorithmic fairness over time.
\emph{Explainability} is crucial not only for compliance and fostering trust~\citep{schmude2023impact} but may also prevent unexpected and unfavorable reactions from affected individuals who lack understanding of decisions~\citep{dodge2019explaining, schoffer2022relationship}. These reactions can impact the system dynamics, requiring further research on the temporal relationship between explanations and fairness.
\emph{Algorithmic recourse} provides individuals who received negative decisions at one time step with recommended feature changes to receive positive decisions in the next step, thereby also affecting the dynamics of a system. While fair recourse has been studied~\citep{heidari2019effort, von2020fairness, kavouras2024fairness, gupta2019equalizing}, open questions regarding the long-term fairness and impact of recourse persist, including ensuring fairness when the same entity (e.g., a bank) controls both decision-making and recourse policies.
Finally, as large language models (\LLMs) are increasingly deployed in society~\citep{shumailov2023curse} and used for tabular decision making~\citep{gallegos2023bias,li2023fairness, liu2023investigating}, studying temporal aspects in \emph{detecting and mitigating \LLM\ harms}~\citep{hadi2023survey, cui2023fft} becomes important~\cite{susser2022decision}. 
Future research should explore and develop a deeper understanding of temporal fairness challenges in \LLMs.

\section*{Acknowledgments}
Thank you very much to Diego Baptista Theuerkauf for helpful comments and suggestions. MR appreciates the generous financial support by the 2023 Google PhD Fellowship in Machine Learning.

\bibliography{references}

@article{pagan2023classification,
  title={A Classification of Feedback Loops and Their Relation to Biases in Automated Decision-Making Systems},
  author={Pagan, Nicol{\`o} and Baumann, Joachim and Elokda, Ezzat and De Pasquale, Giulia and Bolognani, Saverio and Hann{\'a}k, Anik{\'o}},
  journal={arXiv preprint arXiv:2305.06055},
  year={2023}
}

@inproceedings{heidari2018preventing,
  title={Preventing Disparate Treatment in Sequential Decision Making.},
  author={Heidari, Hoda and Krause, Andreas},
  booktitle={IJCAI},
  pages={2248--2254},
  year={2018}
}

@article{gupta2019equalizing,
  title={Equalizing recourse across groups},
  author={Gupta, Vivek and Nokhiz, Pegah and Roy, Chitradeep Dutta and Venkatasubramanian, Suresh},
  journal={arXiv preprint arXiv:1909.03166},
  year={2019}
}

@article{kavouras2024fairness,
  title={Fairness Aware Counterfactuals for Subgroups},
  author={Kavouras, Loukas and Tsopelas, Konstantinos and Giannopoulos, Giorgos and Sacharidis, Dimitris and Psaroudaki, Eleni and Theologitis, Nikolaos and Rontogiannis, Dimitrios and Fotakis, Dimitris and Emiris, Ioannis},
  journal={Advances in Neural Information Processing Systems},
  volume={36},
  year={2024}
}

@inbook{kirchhefer2023algorithmicgroups, series={Digital | Recht}, 
author = {Kirchhefer-Lauber, Anna},
title={Der Schutz von algorithmischen Gruppen im deutschen Nichtdiskriminierungsrecht}, year={2023}}

@article{sen2016race,
  title={Race as a bundle of sticks: Designs that estimate effects of seemingly immutable characteristics},
  author={Sen, Maya and Wasow, Omar},
  journal={Annual Review of Political Science},
  volume={19},
  pages={499--522},
  year={2016},
  publisher={Annual Reviews}
}

@inproceedings{schwobel2022long,
  title={The long arc of fairness: Formalisations and ethical discourse},
  author={Schw{\"o}bel, Pola and Remmers, Peter},
  booktitle={Proceedings of the 2022 ACM Conference on Fairness, Accountability, and Transparency},
  pages={2179--2188},
  year={2022}
}

@inproceedings{kumar2019credit,
  title={Credit card fraud detection using random forest algorithm},
  author={Kumar, M Suresh and Soundarya, V and Kavitha, S and Keerthika, ES and Aswini, E},
  booktitle={2019 3rd International Conference on Computing and Communications Technologies (ICCCT)},
  pages={149--153},
  year={2019},
  organization={IEEE}
}

@article{lum2016statistical,
  title={A statistical framework for fair predictive algorithms},
  author={Lum, Kristian and Johndrow, James},
  journal={arXiv preprint arXiv:1610.08077},
  year={2016}
}

@article{shumailov2023curse,
  title={The Curse of Recursion: Training on Generated Data Makes Models Forget.(2023)},
  author={Shumailov, Ilia and Shumaylov, Zakhar and Zhao, Yiren and Gal, Yarin and Papernot, Nicolas and Anderson, Ross},
  journal={arXiv preprint arxiv:2305.17493},
  year={2023}
}

@inproceedings{liu2023investigating,
  title={Investigating the Fairness of Large Language Models for Predictions on Tabular Data},
  author={Liu, Yanchen and Gautam, Srishti and Ma, Jiaqi and Lakkaraju, Himabindu},
  booktitle={Socially Responsible Language Modelling Research},
  year={2023}
}

@inproceedings{yin2023long,
	year = {2023},
	booktitle = {ICLR 2023 Workshop on Trustworthy and Reliable Large-Scale Machine Learning Models},
	author = {Yin, Tongxin and Raab, Reilly and Liu, Mingyan and Liu, Yang},
	title = {Long-Term Fairness with Unknown Dynamics},
}

@inproceedings{wen2021algorithms,
	organization = {PMLR},
	year = {2021},
	pages = {1144--1152},
	booktitle = {Proceedings of the International Conference on Artificial Intelligence and Statistics (AISTATS)},
	author = {Wen, Min and Bastani, Osbert and Topcu, Ufuk},
	title = {Algorithms for fairness in sequential decision making},
}

@inproceedings{chi2022towards,
	organization = {PMLR},
	year = {2022},
	pages = {1161--1178},
	booktitle = {Proceedings of the International Conference on Artificial Intelligence and Statistics (AISTATS)},
	author = {Chi, Jianfeng and Shen, Jian and Dai, Xinyi and Zhang, Weinan and Tian, Yuan and Zhao, Han},
	title = {Towards Return Parity in Markov Decision Processes},
}

@article{yu2022policy,
	year = {2022},
	pages = {8211--8213},
	volume = {35},
	journal = {Advances in Neural Information Processing Systems (NeurIPS)},
	author = {Yu, Eric and Qin, Zhizhen and Lee, Min Kyung and Gao, Sicun},
	title = {Policy Optimization with Advantage Regularization for Long-Term Fairness in Decision Systems},
}

@inproceedings{chaney2018algorithmic,
	year = {2018},
	pages = {224--232},
	booktitle = {Proceedings of the 12th ACM conference on recommender systems},
	author = {Chaney, Allison JB and Stewart, Brandon M and Engelhardt, Barbara E},
	title = {How algorithmic confounding in recommendation systems increases homogeneity and decreases utility},
}

@article{fuster2022predictably,
	publisher = {Wiley Online Library},
	year = {2022},
	pages = {5--47},
	number = {1},
	volume = {77},
	journal = {The Journal of Finance},
	author = {Fuster, Andreas and Goldsmith-Pinkham, Paul and Ramadorai, Tarun and Walther, Ansgar},
	title = {Predictably unequal? The effects of machine learning on credit markets},
}

@inproceedings{mhasawade2021causal,
	year = {2021},
	pages = {784--794},
	booktitle = {Proceedings of the 2021 AAAI/ACM Conference on AI, Ethics, and Society},
	author = {Mhasawade, Vishwali and Chunara, Rumi},
	title = {Causal multi-level fairness},
}

@inproceedings{creager2020causal,
	organization = {PMLR},
	year = {2020},
	pages = {2185--2195},
	booktitle = {International Conference on Machine Learning},
	author = {Creager, Elliot and Madras, David and Pitassi, Toniann and Zemel, Richard},
	title = {Causal modeling for fairness in dynamical systems},
}

@article{gallegos2023bias,
  title={Bias and fairness in large language models: A survey},
  author={Gallegos, Isabel O and Rossi, Ryan A and Barrow, Joe and Tanjim, Md Mehrab and Kim, Sungchul and Dernoncourt, Franck and Yu, Tong and Zhang, Ruiyi and Ahmed, Nesreen K},
  journal={arXiv preprint arXiv:2309.00770},
  year={2023}
}

@article{hadi2023survey,
  title={A survey on large language models: Applications, challenges, limitations, and practical usage},
  author={Hadi, Muhammad Usman and Qureshi, Rizwan and Shah, Abbas and Irfan, Muhammad and Zafar, Anas and Shaikh, Muhammad Bilal and Akhtar, Naveed and Wu, Jia and Mirjalili, Seyedali and others},
  journal={Authorea Preprints},
  year={2023},
  publisher={Authorea}
}

@article{cui2023fft,
  title={FFT: Towards Harmlessness Evaluation and Analysis for LLMs with Factuality, Fairness, Toxicity},
  author={Cui, Shiyao and Zhang, Zhenyu and Chen, Yilong and Zhang, Wenyuan and Liu, Tianyun and Wang, Siqi and Liu, Tingwen},
  journal={arXiv preprint arXiv:2311.18580},
  year={2023}
}

@article{li2023fairness,
  title={Fairness of ChatGPT},
  author={Li, Yunqi and Zhang, Yongfeng},
  journal={arXiv preprint arXiv:2305.18569},
  year={2023}
}

@inproceedings{agarwal2018reductions,
	year = {2018},
	pages = {60--69},
	booktitle = {Proceedings of the 35th International Conference on Machine Learning, ICML 2018},
	author = {Agarwal, Alekh and Beygelzimer, Alina and Dudik, Miroslav and Langford, John and Wallach, Hanna},
	title = {A Reductions Approach to Fair Classification},
}

@inproceedings{zafar2017fairness,
	year = {2017},
	pages = {1171--1180},
	booktitle = {Proceedings of the 26th International Conference on World Wide Web},
	author = {Zafar, Muhammad Bilal and Valera, Isabel and Gomez Rodriguez, Manuel and Gummadi, Krishna P},
	title = {Fairness beyond disparate treatment \& disparate impact: Learning classification without disparate mistreatment},
}

@misc{COMPAS,
	note = {https://github.com/propublica/compas-analysis [Online; Accessed 22.06.2020]},
	title = {ProPublica COMPAS analysis and dataset},
	year = {2016},
	author = {Larson, Jeff and 
Atlidakis, Vaggelis and 
Roswell, Marjorie},
}

@inproceedings{liu2018delayed,
	year = {2018},
	booktitle = {Proceedings of the 35th International Conference on Machine Learning, ICML 2018},
	author = {Liu, Lydia T and Dean, Sarah and Rolf, Esther and Simchowitz, Max and Hardt, Moritz},
	title = {Delayed Impact of Fair Machine Learning},
}

@inproceedings{perdomo2020performative,
	organization = {PMLR},
	year = {2020},
	pages = {7599--7609},
	booktitle = {International Conference on Machine Learning},
	author = {Perdomo, Juan and Zrnic, Tijana and Mendler-D{\"u}nner, Celestine and Hardt, Moritz},
	title = {Performative prediction},
}

@inproceedings{heidari2019effort,
	organization = {PMLR},
	year = {2019},
	pages = {2692--2701},
	booktitle = {International Conference on Machine Learning},
	author = {Heidari, Hoda and Nanda, Vedant and Gummadi, Krishna},
	title = {On the Long-term Impact of Algorithmic Decision Policies: Effort Unfairness and Feature Segregation through Social Learning},
}

@inproceedings{dwork2012fairness,
	year = {2012},
	pages = {214--226},
	booktitle = {Proceedings of the 3rd innovations in theoretical computer science conference},
	author = {Dwork, Cynthia and Hardt, Moritz and Pitassi, Toniann and Reingold, Omer and Zemel, Richard},
	title = {Fairness through awareness},
}

@article{hardt2016equality,
	year = {2016},
	volume = {29},
	journal = {Advances in Neural Information Processing Systems (NeurIPS)},
	author = {Hardt, Moritz and Price, Eric and Srebro, Nati},
	title = {Equality of opportunity in supervised learning},
}

@article{xu2023equal,
	year = {2023},
	journal = {arXiv preprint arXiv:2309.03426},
	author = {Xu, Yuancheng and Deng, Chenghao and Sun, Yanchao and Zheng, Ruijie and Wang, Xiyao and Zhao, Jieyu and Huang, Furong},
	title = {Equal Long-term Benefit Rate: Adapting Static Fairness Notions to Sequential Decision Making},
}

@inproceedings{von2020fairness,
	year = {2022},
	pages = {9584--9594},
	volume = {36},
	booktitle = {Proceedings of the AAAI Conference on Artificial Intelligence},
	author = {von K{\"u}gelgen, Julius and Karimi, Amir-Hossein and Bhatt, Umang and Valera, Isabel and Weller, Adrian and Sch{\"o}lkopf, Bernhard},
	title = {On the fairness of causal algorithmic recourse},
}

@inproceedings{hu2020sex,
	year = {2020},
	pages = {513--513},
	booktitle = {Proceedings of the Conference on Fairness, Accountability, and Transparency (FAccT)},
	author = {Hu, Lily and Kohler-Hausmann, Issa},
	title = {What's sex got to do with machine learning?},
}

@article{wachter2022theory,
	publisher = {HeinOnline},
	year = {2022},
	pages = {149},
	volume = {97},
	journal = {Tul. L. Rev.},
	author = {Wachter, Sandra},
	title = {The theory of artificial immutability: Protecting algorithmic groups under anti-discrimination law},
}

@inproceedings{kilbertus2020fair,
	organization = {PMLR},
	year = {2020},
	pages = {277--287},
	booktitle = {Proceedings of the International Conference on Artificial Intelligence and Statistics (AISTATS)},
	author = {Kilbertus, Niki and Rodriguez, Manuel Gomez and Sch{\"o}lkopf, Bernhard and Muandet, Krikamol and Valera, Isabel},
	title = {Fair decisions despite imperfect predictions},
}

@inproceedings{jabbari2017fairness,
	organization = {PMLR},
	year = {2017},
	pages = {1617--1626},
	booktitle = {International conference on machine learning},
	author = {Jabbari, Shahin and Joseph, Matthew and Kearns, Michael and Morgenstern, Jamie and Roth, Aaron},
	title = {Fairness in reinforcement learning},
}

@article{schoffer2022relationship,
  title={On the relationship between explanations, fairness perceptions, and decisions},
  author={Sch{\"o}ffer, Jakob and De-Arteaga, Maria and Kuehl, Niklas},
  journal={ACM CHI 2022 Workshop on Human-Centered Explainable AI (HCXAI)},
  year={2022}
}

@article{gupta2023temporal,
  title={Temporal Fairness in Online Decision-Making},
  author={Gupta, Swati and Kamble, Vijay and Salem, Jad},
  journal={Ethics in Artificial Intelligence: Bias, Fairness and Beyond},
  pages={53--69},
  year={2023},
  publisher={Springer}
}

@inproceedings{dodge2019explaining,
  title={Explaining models: an empirical study of how explanations impact fairness judgment},
  author={Dodge, Jonathan and Liao, Q Vera and Zhang, Yunfeng and Bellamy, Rachel KE and Dugan, Casey},
  booktitle={Proceedings of the 24th international conference on intelligent user interfaces},
  pages={275--285},
  year={2019}
}

@inproceedings{schumann2020we,
  title={We need fairness and explainability in algorithmic hiring},
  author={Schumann, Candice and Foster, Jeffrey and Mattei, Nicholas and Dickerson, John},
  booktitle={International Conference on Autonomous Agents and Multi-Agent Systems (AAMAS)},
  year={2020}
}

@inproceedings{schmude2023impact,
  title={On the Impact of Explanations on Understanding of Algorithmic Decision-Making},
  author={Schmude, Timoth{\'e}e and Koesten, Laura and M{\"o}ller, Torsten and Tschiatschek, Sebastian},
  booktitle={Proceedings of the 2023 ACM Conference on Fairness, Accountability, and Transparency},
  pages={959--970},
  year={2023}
}

@article{sachs2005causal,
	publisher = {American Association for the Advancement of Science},
	year = {2005},
	pages = {523--529},
	number = {5721},
	volume = {308},
	journal = {Science},
	author = {Sachs, Karen and Perez, Omar and Pe'er, Dana and Lauffenburger, Douglas A and Nolan, Garry P},
	title = {Causal protein-signaling networks derived from multiparameter single-cell data},
}

@article{vowels2021d,
	publisher = {ACM New York, NY},
	year = {2022},
	pages = {1--36},
	number = {4},
	volume = {55},
	journal = {ACM Computing Surveys (CSUR)},
	author = {Vowels, Matthew J and Camgoz, Necati Cihan and Bowden, Richard},
	title = {D’ya like dags? a survey on structure learning and causal discovery},
}

@article{wachter2020bias,
	publisher = {HeinOnline},
	year = {2020},
	pages = {735},
	volume = {123},
	journal = {W. Va. L. Rev.},
	author = {Wachter, Sandra and Mittelstadt, Brent and Russell, Chris},
	title = {Bias preservation in machine learning: the legality of fairness metrics under EU non-discrimination law},
}

@article{zhang2020fair,
	year = {2020},
	pages = {18457--18469},
	volume = {33},
	journal = {Advances in Neural Information Processing Systems},
	author = {Zhang, Xueru and Tu, Ruibo and Liu, Yang and Liu, Mingyan and Kjellstrom, Hedvig and Zhang, Kun and Zhang, Cheng},
	title = {How do fair decisions fare in long-term qualification?},
}

@article{chouldechova2018frontiers,
	year = {2018},
	journal = {arXiv preprint arXiv:1810.08810},
	author = {Chouldechova, Alexandra and Roth, Aaron},
	title = {The frontiers of fairness in machine learning},
}

@article{puranikdynamic,
	year = {2022},
	journal = {ICLR Workshop on Socially Responsible Machine Learning},
	author = {Puranik, Bhagyashree and Madhow, Upamanyu and Pedarsani, Ramtin},
	title = {DYNAMIC POSITIVE REINFORCEMENT FOR LONG-TERM FAIRNESS},
}

@inproceedings{lakkaraju2017selective,
	year = {2017},
	pages = {275--284},
	booktitle = {Proceedings of the 23rd ACM SIGKDD International Conference on Knowledge Discovery and Data Mining},
	author = {Lakkaraju, Himabindu and Kleinberg, Jon and Leskovec, Jure and Ludwig, Jens and Mullainathan, Sendhil},
	title = {The selective labels problem: Evaluating algorithmic predictions in the presence of unobservables},
}

@article{bechavod2019equal,
	year = {2019},
	volume = {32},
	journal = {Advances in Neural Information Processing Systems},
	author = {Bechavod, Yahav and Ligett, Katrina and Roth, Aaron and Waggoner, Bo and Wu, Steven Z},
	title = {Equal opportunity in online classification with partial feedback},
}

@article{hardt2022performative,
  title={Performative power},
  author={Hardt, Moritz and Jagadeesan, Meena and Mendler-D{\"u}nner, Celestine},
  journal={Advances in Neural Information Processing Systems},
  volume={35},
  pages={22969--22981},
  year={2022}
}

@inproceedings{chiappa2019path,
	year = {2019},
	pages = {7801--7808},
	volume = {33},
	booktitle = {Proceedings of the AAAI Conference on Artificial Intelligence},
	author = {Chiappa, Silvia},
	title = {Path-specific counterfactual fairness},
}

@article{zafar2019fairness,
	url = {http://jmlr.org/papers/v20/18-262.html},
	pages = {1-42},
	number = {75},
	volume = {20},
	year = {2019},
	journal = {Journal of Machine Learning Research},
	title = { Fairness Constraints: A Flexible Approach for Fair Classification },
	author = {Muhammad Bilal Zafar and Isabel Valera and Manuel Gomez-Rodriguez and Krishna P. Gummadi},
}

@inproceedings{zafar2017bfairness,
	url = {https://proceedings.mlr.press/v54/zafar17a.html},
	publisher = {PMLR},
	month = {20--22 Apr},
	series = {Proceedings of Machine Learning Research},
	volume = {54},
	editor = {Singh, Aarti and Zhu, Jerry},
	year = {2017},
	pages = {962--970},
	booktitle = {Proceedings of the International Conference on Artificial Intelligence and Statistics (AISTATS)},
	author = {Zafar, Muhammad Bilal and Valera, Isabel and Rogriguez, Manuel Gomez and Gummadi, Krishna P.},
	title = {{Fairness Constraints: Mechanisms for Fair Classification}},
}

@article{kilbertus2017avoiding,
	year = {2017},
	volume = {30},
	journal = {Advances in neural information processing systems},
	author = {Kilbertus, Niki and Rojas Carulla, Mateo and Parascandolo, Giambattista and Hardt, Moritz and Janzing, Dominik and Sch{\"o}lkopf, Bernhard},
	title = {Avoiding discrimination through causal reasoning},
}

@inproceedings{damour2020fairness,
	year = {2020},
	pages = {525--534},
	booktitle = {Proceedings of the Conference on Fairness, Accountability, and Transparency (FAccT)},
	author = {D'Amour, Alexander and Srinivasan, Hansa and Atwood, James and Baljekar, Pallavi and Sculley, D and Halpern, Yoni},
	title = {Fairness is not static: deeper understanding of long term fairness via simulation studies},
}

@article{gupta2021individual,
	year = {2021},
	pages = {1--35},
	number = {144},
	volume = {22},
	journal = {J. Mach. Learn. Res.},
	author = {Gupta, Swati and Kamble, Vijay},
	title = {Individual Fairness in Hindsight.},
}

@inproceedings{ghassami2020characterizing,
	organization = {PMLR},
	year = {2020},
	pages = {3494--3504},
	booktitle = {International Conference on Machine Learning},
	author = {Ghassami, AmirEmad and Yang, Alan and Kiyavash, Negar and Zhang, Kun},
	title = {Characterizing distribution equivalence and structure learning for cyclic and acyclic directed graphs},
}

@article{strobl2019constraint,
	publisher = {Springer},
	year = {2019},
	pages = {33--56},
	number = {1},
	volume = {8},
	journal = {International Journal of Data Science and Analytics},
	author = {Strobl, Eric V},
	title = {A constraint-based algorithm for causal discovery with cycles, latent variables and selection bias},
}

@phdthesis{kilbertus2021beyond,
  title={Beyond traditional assumptions in fair machine learning},
  author={Kilbertus, Niki},
  year={2020},
  school={University of Cambridge}
}

@inproceedings{susser2022decision,
  title={Decision Time: Normative Dimensions of Algorithmic Speed},
  author={Susser, Daniel},
  booktitle={Proceedings of the 2022 ACM Conference on Fairness, Accountability, and Transparency},
  pages={1410--1420},
  year={2022}
}

@article{Hu_Zhang_2022,
	pages = {9549-9557},
	year = {2022},
	author = {Hu, Yaowei and Zhang, Lu},
	journal = {Proceedings of the AAAI Conference on Artificial Intelligence},
	number = {9},
	volume = {36},
	title = {Achieving Long-Term Fairness in Sequential Decision Making},
}

@article{alamdari2023remembering,
  title={Remembering to Be Fair: On Non-Markovian Fairness in Sequential DecisionMaking (Preliminary Report)},
  author={Alamdari, Parand A and Klassen, Toryn Q and Creager, Elliot and McIlraith, Sheila A},
  journal={arXiv preprint arXiv:2312.04772},
  year={2023}
}

@article{scantamburlo2023prediction,
  title={On Prediction-Modelers and Decision-Makers: Why Fairness Requires More Than a Fair Prediction Model},
  author={Scantamburlo, Teresa and Baumann, Joachim and Heitz, Christoph},
  journal={arXiv preprint arXiv:2310.05598},
  year={2023}
}

@article{zezulka2023fairness,
  title={Fairness After Intervention: Towards a Theory of Substantial Fairness for Machine Learning},
  author={Zezulka, Sebastian},
  year={2023}
}

@article{weber2022enforcing,
  title={Enforcing delayed-impact fairness guarantees},
  author={Weber, Aline and Metevier, Blossom and Brun, Yuriy and Thomas, Philip S and da Silva, Bruno Castro},
  journal={arXiv preprint arXiv:2208.11744},
  year={2022}
}

@article{zhao2023fairo,
  title={FAIRO: Fairness-aware Adaptation in Sequential-Decision Making for Human-in-the-Loop Systems},
  author={Zhao, Tianyu and Taherisadr, Mojtaba and Elmalaki, Salma},
  journal={arXiv preprint arXiv:2307.05857},
  year={2023}
}

@inproceedings{bohdal2023fairness,
  title={Fairness in AI and Its Long-Term Implications on Society},
  author={Bohdal, Ondrej and Hospedales, Timothy and Torr, Philip HS and Barez, Fazl},
  booktitle={Third Annual Stanford Existential Risks Conference},
  year={2023}
}

@article{rateike2023designing,
  title={Designing Long-term Group Fair Policies in Dynamical Systems},
  author={Rateike, Miriam and Valera, Isabel and Forr{\'e}, Patrick},
  journal={arXiv preprint arXiv:2311.12447},
  year={2023}
}

@inproceedings{paassen2019dynamic,
  title={Dynamic fairness-Breaking vicious cycles in automatic decision making},
  author={Paa{\ss}en, Benjamin and Bunge, Astrid and Hainke, Carolin and Sindelar, Leon and Vogelsang, Matthias},
  booktitle={Proceedings of the 27th European Symposium on Artificial Neural Networks (ESANN 2019)},
  year={2019}
}

@article{Rateike2022DontTI,
	year = {2022},
	journal = {2022 ACM Conference on Fairness, Accountability, and Transparency},
	author = {Miriam Rateike and Ayan Majumdar and Olga Mineeva and Krishna P. Gummadi and Isabel Valera},
	title = {Don’t Throw it Away! The Utility of Unlabeled Data in Fair Decision Making},
}

\end{document}